\documentclass[a4paper,fleqn]{cas-dc}
\usepackage[numbers,sort&compress]{natbib}
\usepackage{multirow}
\usepackage{dsfont}
\usepackage{placeins} 
\usepackage[pagewise]{lineno}

\def\tsc#1{\csdef{#1}{\textsc{\lowercase{#1}}\xspace}}
\tsc{WGM}
\tsc{QE}
\tsc{EP}
\tsc{PMS}
\tsc{BEC}
\tsc{DE}

\usepackage{tabularx}
\begin{document}
\let\WriteBookmarks\relax
\def\floatpagepagefraction{1}
\def\textpagefraction{.001}

\shorttitle{OSMLoc}

\shortauthors{Liao et~al.}

\title [mode = title]{OSMLoc: Single Image-Based Visual Localization in OpenStreetMap with Fused Geometric and Semantic Guidance}                      

\author[1]{Youqi Liao}

\cormark[1]

\affiliation[1]{organization={Wuhan University}, 
    city={Wuhan},
    postcode={430079}, 
    country={China}}

\author[2]{Xieyuanli Chen} 
\cormark[1]

\affiliation[2]{organization={National University of Defense Techology},
    city={Changsha},
    postcode={410003}, 
    country={China}}

\author%
[3]
{Shuhao Kang}

\affiliation[3]{organization={Technical University of Munich},
    city={Munich},
    postcode={D-80333}, 
    country={Germany}}

\author[4]{Jianping Li}
\cormark[2]

\affiliation[4]{organization={Nanyang Technological University},
    city={Singapore},
    postcode={639798}, 
    country={Singapore}}

\affiliation[5]{organization={Norwegian University of Science and Technology},
    city={Trondheim},
    postcode={NO-7491 }, 
    country={Norwegian}}

\author[1]{Zhen Dong}
\author[5]{Hongchao Fan}
\author[1]{Bisheng Yang}

\cortext[cor1]{Equal Contribution}
\cortext[cor2]{Corresponding author}
\tnotetext[1]{This research was supported by the National Natural Science Foundation Project ( No. 42201477, No.42130105)} 

\renewcommand{\arraystretch}{1.05}
\begin{abstract}
 OpenStreetMap (OSM), a rich and versatile source of volunteered geographic information (VGI), facilitates human self-localization and scene understanding by integrating nearby visual observations with vectorized map data. However, the disparity in modalities and perspectives poses a major challenge for effectively matching camera imagery with compact map representations, thereby limiting the full potential of VGI data in real-world localization applications.
 Inspired by the fact that the human brain relies on the fusion of geometric and semantic understanding for spatial localization tasks~\citep{hermer1994geometric,epstein2014neural}, we propose the OSMLoc in this paper. OSMLoc is a brain-inspired visual localization approach based on first-person-view images against the OSM maps. It integrates semantic and geometric guidance to significantly improve accuracy, robustness, and generalization capability. First, we equip the OSMLoc with the visual foundational model to extract powerful image features. Second, a geometry-guided depth distribution adapter is proposed to bridge the monocular depth estimation and camera-to-BEV transform. Thirdly, the semantic embeddings from the OSM data are utilized as auxiliary guidance for image-to-OSM feature matching. To validate the proposed OSMLoc, we collect a worldwide cross-area and cross-condition (CC) benchmark for extensive evaluation. Experiments on the MGL dataset, CC validation benchmark, and KITTI dataset have demonstrated the superiority of our method. Code, pre-trained models, CC validation benchmark, and additional results are available at: \url{https://github.com/WHU-USI3DV/OSMLoc}.
\end{abstract}

\begin{highlights}
\item OSMLoc: a brain-inspired framework jointly fusing geometry and semantics for I2O localization.
\item  Geometry-guided BEV transform with foundation model's depth prior.
\item Auxiliary semantic alignment task boosts image model's scene understanding.
\item Cross-area \& cross-condition benchmark for extensive evaluation.
\end{highlights}

\begin{keywords}
Visual Localization \sep  
OpenStreetMap \sep 
Visual Foundational Model \sep
Cross-modality Fusion
\end{keywords}

\maketitle

\section{Introduction}
\switchlinenumbers
The fusion of visual observation with geo-referenced maps forms the foundation for seamless navigation in logistic robots and unmanned ground vehicles (UGV)~\citep{yuan2017cooperative,zhu2021vigor,yang2023uplp, zhou2025learning}. Most existing industrial solutions use the geo-referenced images~\citep{arandjelovic2016netvlad,piras2017information,hausler2021patch,li2021image} or pre-built 3D point cloud maps ~\citep{yu2021deep,chen2022i2d,shi2022improved,zou2023patchaugnet} as the reference map. However, constructing geo-referenced images or point cloud maps is expensive and difficult to update. 
Since the beginning of the Internet era, volunteering geo-informatics (VGI) has emerged as a new paradigm that creates, assembles, and disseminates geographic data by crowd-sourcing users instead of specific experts ~\citep{wang2024framework}. 
Among VGI projects, OpenStreetMap (OSM) stands out as one of the most successful collaborative mapping initiatives, with over 10 million registered contributors~\citep{OSM2024}. OSM provides detailed and up-to-date data about stationary objects throughout the world, including infrastructure and other aspects of the built environment, points of interest, land use and land cover classifications, and topographical features~\citep{fan2014quality,vargas2020openstreetmap}. 
With its rich fusion of geographic and semantic content, humans can easily localize themselves by comparing the surrounding views with the OSM web maps. However, the image-to-OSM (I2O) localization process remains challenging for autonomous systems due to cross-modal discrepancies and drastic viewpoint variations, which limit the full potential of the OSM data in the logistics industry and hinder the development of navigation with the crowd-sourced map.

Due to above mentioned difficulties, research on I2O visual localization is scarce. Early I2O localization approaches, such as \citet{samano2020you} localize along the pre-defined routes, while \citet{zhou2021efficient} integrates deep image features into the Monte Carlo framework to improve the representation ability of the observation model. OrienterNet~\citep{sarlin2023orienternet} is the first end-to-end single-image I2O localization approach, which lifts the front-view image features to the birds-eye-view (BEV) and estimates the pose by dense matching. MaplocNet~\citep{wu2024maplocnet} proposed a coarse-to-fine registration pipeline for I2O localization and introduced the semantic labels from the autonomous driving datasets as auxiliary supervision. Although existing methods have achieved impressive results, we highlight that two key challenges remain unsolved: 
1)  effective modular transformation of visual observations for cross-modality  fusion and I2O localization;
2)  explainable localization and strong generalization across diverse environments worldwide.

\begin{figure}[!t]
\centering
\includegraphics[width=3.3in]{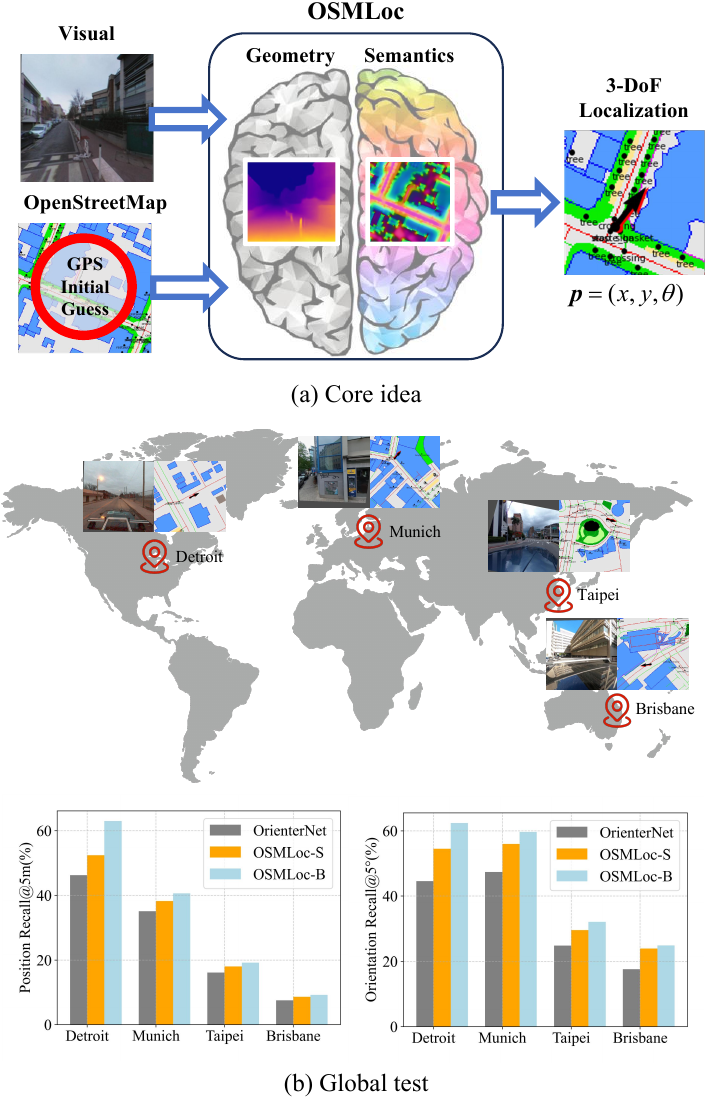}
\caption{Image-to-OpenstreetMap localization aims to estimate the 3 degrees of freedom (3-DoF) camera pose. (a) shows the core idea of our method, which fuses geometric and semantic guidance into the framework.  (b) shows the global evaluation results.}
\label{fig:motivation}
\end{figure}

Fortunately, the human navigation system provides profound inspiration for addressing these challenges. \citet{hermer1994geometric} first demonstrated that even young children can utilize geometric cues from their surroundings for orientation and localization. Building on this, \citep{epstein2014neural,epstein2017cognitive} showed that the retrosplenial–hippocampal circuit in the brain supports robust localization through the integration of geometric and semantic information, a principle that directly motivates our framework. Further behavioral evidence from \citep{foo2005humans} indicates that humans construct cognitive maps by combining environmental geometry with landmark semantics. Building upon the foundation, we introduce the OSMLoc, a single image-based visual localization framework with fused geometric and semantic guidance.  The core idea of OSMLoc is illustrated in Fig.~\ref{fig:motivation}(a).
We inherit the fundamental model with rich prior knowledge, specifically, the DINOv2 encoder~\citep{oquab2023dinov2} of the Depth Anything model~\citep{yang2024depth} into our framework to construct powerful image features. A compact depth distribution adapter is proposed to incorporate geometric information from Depth Anything into camera-to-BEV transformation. To our knowledge, this is the first attempt to integrate the foundation model into the I2O visual localization framework. The semantic consistency between visual and OSM data aids the model's learning in completing cross-modular matching.
Moreover, existing same-area validation sets exhibit similar styles, with consistent building and road patterns within neighborhood blocks of the training sets. This similarity arises from the fact the data is typically collected by the same users, on similar devices, and within close timeframes. Such uniformity may introduce bias into the validation results. Therefore, we collect a novel cross-area and cross-condition (CC) validation benchmark for evaluation in diverse environments, as shown in Fig.~\ref{fig:motivation}(b). The CC validation benchmark contains thousands of images from Detroit, Munich, Taipei, and Brisbane with vast scenarios.
Experiments on the Mapillary geo-localization (MGL) dataset, CC validation benchmark, and the KITTI dataset reveal that our method outperforms existing methods, particularly in unseen areas. 
Overall, the main contributions include:
\begin{enumerate}
    \item We propose a brain-inspired framework, OSMLoc, which jointly learns the geometry, semantics, and pose for I2O visual localization. Given the challenges of localization across cross-view and cross-modality sources, we fully leverage geometric and semantic cues from the foundational model and OSM data to help the framework bridge the domain gap and understand the surroundings. 
    \item We propose a novel depth distribution adapter (DDA) to incorporate the depth prior to the camera-to-BEV transformation. Since accurate depth is essential for viewpoint transformation and monocular depth estimation is ambiguous, we utilize the DDA to distill the prior depth knowledge from the foundation model into our framework.
    \item We introduce the auxiliary semantic alignment task to improve the scene understanding capability of the image model in a self-supervision manner. As the OSM data provides detailed semantic and geographic features of the areas, roads and other objects, we leverage the semantic consistency between the image and map as an additional constraint for the visual localization task.
    \item To evaluate generalization, we collect the cross-area and cross-condition (CC) validation benchmark with thousands of frames in Detroit, Munich, Taipei, and Brisbane, providing an extensive testbed for the I2O visual localization models.
\end{enumerate}

\section{Related works}

 Visual localization methods can be broadly categorized by data acquisition type into two groups: visual localization with expert data (Section \ref{section:review_expertdata}) and  VGI (Section \ref{section:review_vgi}). Expert data refers to data collected, processed and validated by professionals or specialized institutions with industrial-grade devices and rigorous methodologies, such as point cloud and aerial images. VGI is always created, assembled, and processed by individuals using consumer-grade devices and open-source platforms, such as street-view images captured by smartphones and manually curated OSM data.

\subsection{Visual localization with expert data}\label{section:review_expertdata}

The visual localization with point cloud has been studied for decades. Given a query image and pre-built point cloud map, the image-to-point cloud (I2P) visual localization task aims to estimate the transform parameters from the camera coordinate system to the point cloud coordinate system. ~\citet{sattler2011fast} proposed a direct I2P matching framework by assigning point features to visual words and constructing the 2D-3D correspondences by linear search.  Due to the significant computational and storage burden of point cloud, \citet{cheng2016data} proposed a point cloud simplification framework based on the K-Cover theory to improve efficiency. Recent approaches~\citep{li20203d,shi2022improved} focus on learning-based methods to improve accuracy and robustness. Although point cloud contains detailed spatial information for localization, the extraction, processing, and storage are costly and require specialized sensors and software, limiting their applicability in consumer-grade scenarios.

With broad coverage and easy accessibility, aerial image-based visual localization has seen rapid progress. Early image-to-aerial (I2A) methods~\citep{lin2013cross,tian2017cross} treat localization as a retrieval task, dividing aerial images into patches and searching for the most similar one in feature space. VIGOR~\citep{zhu2021vigor} noted the limitations of one-to-one retrieval in real-world scenarios and introduced a coarse-to-fine pipeline with a new dataset. SNAP~\citep{sarlin2024snap} further unified ground-to-aerial localization, semantic segmentation, and mapping within a single framework. To reduce reliance on pose labels, \citet{li2024unleashing} proposed an unsupervised approach leveraging unlabeled data. While more compact than 3D maps, aerial images remain costly to obtain, often require paid access, and consume significant storage at high resolution. Additionally, variations across time, seasons, weather, and lighting conditions make long-term localization with aerial imagery particularly challenging.
 
\subsection{Visual localization with VGI}\label{section:review_vgi}

Unlike expert data, VGI data is created, assembled, and shared by individuals.  Visual localization with the geo-referenced ground images from VGI platforms like Mapillary~\citep{warburg2020mapillary} or KartaView~\citep{Kartaview2024} as the primary databases, is a common research area. Image-to-ground images (I2I) visual localization approaches follow a two-step pipeline: coarse location retrieval and precise pose estimation. For the first step, ~\citep{arandjelovic2016netvlad, unar2018visual, ahmed2019content,li2023hot,zhu2023r2former} treat it as a place recognition task, which retrieves the top candidates and re-ranks the top-k candidates to refine results. For pose estimation, matching-based methods~\citep{dusmanu2019d2,sarlin2020superglue} find pixel correspondences and solve for homography, whereas matching-free methods~\citep{kendall2015posenet, liu2020posegan, tang2024transcnnloc} directly regress 6-DoF poses with a neural pose regressor. 

OSM-based visual localization approaches are most related to our method. The pioneering work OpenStreetSLAM~\citep{floros2013openstreetslam} integrates visual odometry with OSM data to mitigate drift accumulated during vehicle motion. It extracts road line segments from the OSM tiles and uses them as geo-referenced constraints within a Monte Carlo Localization (MCL) framework. \citet{samano2020you} re-formulated the I2O localization task as a global retrieval problem, embedding both query images and OSM map tiles into global descriptors for matching along predefined routes. Building on this formulation, \citet{zhou2021efficient} incorporated these deep image/map features into an MCL framework to enable sequential localization. However, such methods are typically constrained to operate along predefined routes or limited areas, which restricts their applicability in large-scale, open-world scenarios. In addition, their localization accuracy remains relatively low due to the limited representation quality of learned features. More recently, \citet{dong2024augmenting} and \citet{zhou2025seglocnet} combined camera images with LiDAR/Radar point clouds as visual observation to enhance the 3D geometric information. Other approaches~\citep {yan2019global,cho2022openstreetmap,lee2024autonomous,kang2025opal} utilize point clouds as the query for OSM-based localization. OrienterNet~\citep{sarlin2023orienternet} is the first end-to-end, single-image I2O localization framework that lifts image features into the BEV space and estimates camera poses via feature matching. However, its monocular depth estimation for view transformation remains challenging, as the depth associated with each camera pixel is inherently ambiguous~\citep{liu2023bevfusion}. This limits both localization accuracy and generalization to unseen devices or scenes. MapLocNet~\citep{wu2024maplocnet} proposes a multi-task registration pipeline that performs simultaneous pose regression and semantic segmentation. Although it leverages drivable area labels from HD maps and building/POI labels from OSM as supervision, the HD maps are expert-curated and not widely available across diverse regions. Moreover, by ignoring diverse OSM semantic features, such as grass, paths, and road categories, the model overlooks rich geo-spatial context and may yield suboptimal performance. Since MapLocNet~\citep{wu2024maplocnet} relies on drivable area annotations from HD maps, which are expert-curated and not widely available across diverse regions, it requires fine-tuning on downstream datasets to achieve satisfactory performance. To address these limitations, we propose a novel I2O visual localization framework that integrates geometric and semantic cues extracted from foundation models and freely available OSM data. Our approach achieves competitive localization accuracy without relying on manual annotations or task-specific fine-tuning.

\section{Our Approach: OSMLoc}
Given a single image $\mathcal{I}$ with noisy position prior $(\check{x},\check{y}) \in \mathbb{R}^2$, our approach aims to query the precise position and orientation using the OSM map.  As the OSM data mainly consists of 2D geospatial information, we simplify the typical 6-DoF pose estimation problem to the 3-DoF pose $\boldsymbol{p}=(x, y,\theta)$ consisting of the 2-D position $(x, y) \in \mathbb{R}^2$ and orientation angle $\theta \in (-\pi,\pi]$. We assume the gravity direction of the image is accessible via an inertial measurement unit or a gyroscope. The global coordinate system here is the topocentric coordinate system, with the \textit{x-y-z} axes corresponding to \textit{East-North-Up} directions. 

\begin{figure}[!t]
\centering
\includegraphics[width=3.3in]{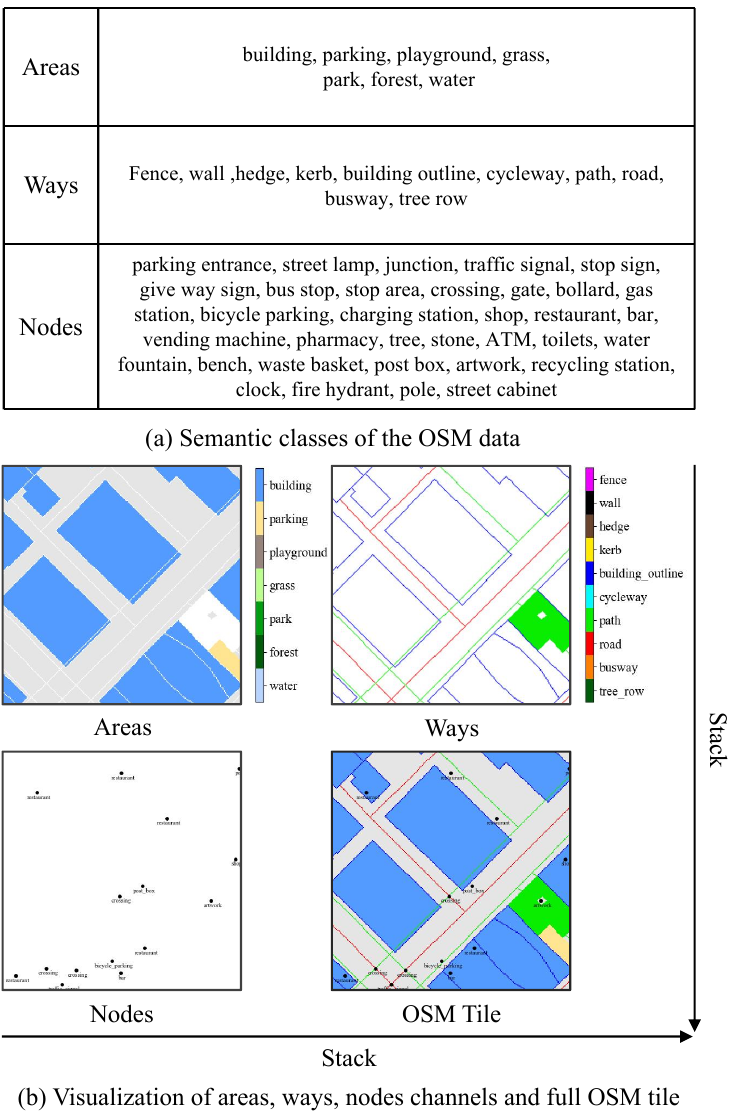}
\caption{Illustration of the OSM pre-processing.}
\label{fig:rasterization}
\end{figure}

Since our approach incorporates cross-view and cross-modality data, the image and OSM data require pre-processing before being fed into the framework. For the image $\mathcal{I}$, we rectify the roll and pitch angles to zero using the known gravity direction and ensure the principal axis is horizontal. The OSM data is stored in structured metadata and represents the areas, ways and nodes with the region of interest (ROI), as well as polylines and point of interest (POI). We categories the elements as in Fig.~\ref{fig:rasterization}(a) and project the OSM data onto the \textit{x-y} plane of the global coordinate system for rasterization. As shown in Fig.~\ref{fig:rasterization}(b), we rasterize the areas, ways and nodes into a 3-channel grid map with a fixed sampling distance $\Delta s = 50\,\text{cm/pixel}$, while preserving the semantic and geographic information. During training and evaluation, we crop map tile $\mathcal{M}$ of the rasterized OSM data centered around the location prior as input. 

\begin{figure}[!t]
\centering
\includegraphics[width=3.3in]{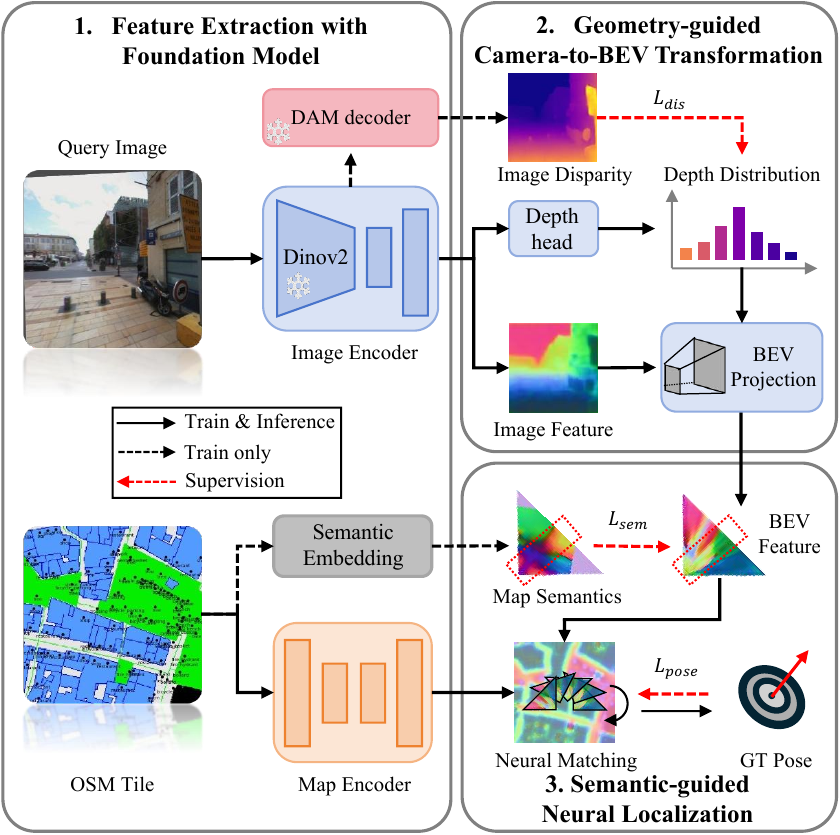}
\caption{Workflow of the OSMLoc.}
\label{fig:workflow}
\end{figure}

The workflow of our method is illustrated in Fig.~\ref{fig:workflow}.  Our OSMLoc first embeds the image $\mathcal{I}$ and OSM map tile $\mathcal{M}$ into high-dimensional feature space (Sec~\ref{sec: fe}), and then the depth distribution adapter distills the geometric prior from foundation model into depth predictions, while the camera-to-BEV transformation module (Sec~\ref{sec:bev}) lifts the image features to birds-eye-view. Lastly, the neural localization (Sec~\ref{sec:loc}) module estimates pose by dense feature matching. The semantic alignment is employed as an auxiliary task to improve the scene understanding ability of the image model. Pose loss $\mathcal{L}_{pose}$, disparity loss $\mathcal{L}_{dis}$ and semantic alignment loss $\mathcal{L}_{sem}$ are leveraged to supervise the network jointly. 

\subsection{Feature extraction with foundation model}\label{sec: fe}

For the image model, we adopt the DINOv2~\cite{oquab2023dinov2} encoder from Depth Anything model (DAM)~\citep{yang2024depth} as our backbone. DINOv2~\citep{oquab2023dinov2} is a large-scale self-supervised vision transformer (ViT)\citep{dosovitskiy2020image} trained on 142\,M diverse images, endowing the model with strong general-purpose visual representations that capture rich semantic concepts and structural cues. Building upon this, DAM fine-tuned DINOv2 on more than 1.5\,M labeled and 62\,M unlabeled images for monocular depth estimation, which further injects geometric priors into the encoder by explicitly learning relative depth relations across scenes.  \\
Crucially, the resulting pre-trained encoder contains not only semantic information but also geometric depth cues, both of which are particularly important for localization. Semantic knowledge facilitates the recognition of place-related entities such as buildings, roads, and vegetation that remain stable across viewpoints, while geometric knowledge improves robustness to perspective changes and enables both view transformation and cross-modal alignment. To preserve these priors, we freeze the pre-trained encoder weights and extract multi-stage features that capture both low-level structures and high-level semantics. The final-layer features are fed into the DAM decoder to predict pseudo ground-truth normalized disparity $\hat{\boldsymbol{t}}$ for geometric-guided view transformation (described in Sec~\ref{sec:bev}). 
As many works~\cite{keetha2023anyloc, mirjalili2023fm} have shown that multi-level features from foundation models benefit localization, we progressively fuse the extracted multi-stage features using a feature pyramid network (FPN) decoder. The resulting image representation $\boldsymbol{F}_{img} \in \mathbb{R}^{U \times V \times C}$ is finally bilinearly upsampled to size $U \times V$ with $C$ channels.

For the map model, we use a multi-layer perception (MLP) to embed the input 3-channel rasterized map $\mathcal{M}$ into a high-dimensional semantic embedding $\boldsymbol{\mathcal{M}}_{sem} \in \mathbb{R}^{H \times W \times (3 \times C_{sem})}$, where $H$ and $W$ are the height and with of the map. Each semantic class of areas, roads and nodes is mapped to a unique semantic vector with $C_{sem}$ channels, respectively. 
Besides, a VGG-19 model is employed as the map encoder, and the decoder is a compact FPN. Similar to the image model, we extract multi-stage features from the encoder and feed them into the FPN to extract the final map feature $\boldsymbol{F}_{map} \in \mathbb{R}^{H \times W \times C}$.

\subsection{Geometry-guided camera-to-BEV transformation}\label{sec:bev}
With the extracted image feature $\boldsymbol{F}_{img} \in \mathbb{R}^{U \times V \times C}$, we infer a BEV feature map $\boldsymbol{F}_{bev} \in \mathbb{R}^{D \times L \times C}$ in a $D \times L$ grid map on the \textit{x-y} plane of the global coordinate system. Following the BEV perception methods~\citep{philion2020lift,liu2023bevfusion}, for each pixel $(u,v)$ on the image, we sample D discrete points $\{(u,d_i,v) \in \mathbb{R}^3 | d_i = d_0 + i\Delta, i \in \{ 1,2,...,D \} \}$ among the camera ray and predict the associated depth distribution $\boldsymbol{\alpha}_{u,v} \in \mathbb{R}^D$. Then we scatter the image feature to the points with the corresponding probability, which generates the feature point cloud $\boldsymbol{F}_{pc} \in \mathbb{R}^{\text{UDV} \times C}$ of size $\text{UDV}$. For each point $(u,d_i,v)$, the associated feature $\boldsymbol{F}_{pc}(u,d_i,v) \in \mathbb{R}^{C}$ is defined as the rescaled feature vector of pixel $(u,v)$:
\begin{equation}
    \boldsymbol{F}_{pc}(u,d_i,v)= \alpha_{u,d_i,v} \boldsymbol{F}_{img}(u,v).
\end{equation} 
With the predicted depth distribution $\boldsymbol{\alpha}_{u,v} \in \mathbb{R}^D$ for each pixel $(u,v)$, the depth value $d_{u,v}$ is formatted as the weighted average value of the depth distribution:
\begin{align}
    d_{u,v} = \sum_{i=1}^D \alpha_{u,d_i,v}d_i, \quad s.t. \sum_{i=1}^D \boldsymbol{\alpha}_{u,d_i,v} = 1.
\end{align}
Since the image is gravity-aligned, each column corresponds to a triangular "slice" in the frustum~\citep{lentsch2023slicematch}, and the associated feature points should lie within the same rectangular region extending from the camera in the BEV polar coordinate system. Therefore, we assemble the feature point cloud of each column into a polar strip on the plane, resulting in a $D \times V$  polar feature map $\boldsymbol{F}_{bev}$. For each polar grid $(d_i,v)$, the feature projection could be formulated as: 

\begin{equation}
    \boldsymbol{F}_{bev}(d_i,v) = \sum_{u=1}^U \alpha_{u,d_i,v} \boldsymbol{F}_{pc}(u,d_i,v),
\end{equation}
where $F_{bev} \in \mathbb{R}^{D \times V \times C}$ is the BEV feature map in the polar coordinate system of the plane. Then we remap the BEV features from the polar coordinate system to a Cartesian grid map of size $D \times L$ by linear interpolation. 

\begin{figure}[ht]
\centering
\includegraphics[width=3.3in]{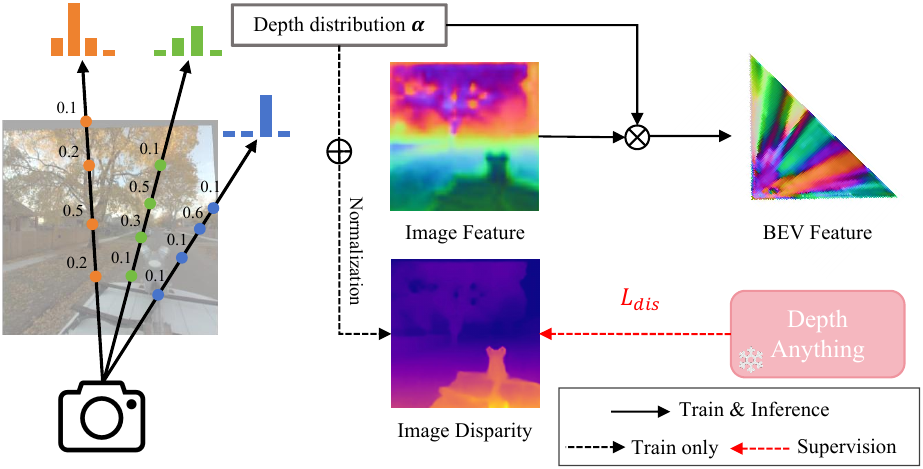}
\caption{Illustration of the depth distribution adapter. For each pixel, we predict the depth distribution $\boldsymbol{\alpha} \in \mathbb{R}^{D}$ and scatter the image feature $\boldsymbol{F}_{img}$ to the BEV feature $\boldsymbol{F}_{bev}$ by the corresponding depth distribution $\boldsymbol{\alpha}$. During training, the depth distribution prediction is transferred to the normalized disparity map and supervised with the Depth Anything.}
\label{fig:c2b}
\end{figure}

\textbf{Depth distribution adapter:}  Although conceptually straightforward, accurate depth distribution is generally unavailable for single-view images in real-world scenarios. External sensors such as LiDAR or RGB-D cameras may provide such information, but relying on them imposes a strong assumption that limits generalization to novel devices and unseen environments. Absolute depth estimators~\citep{yin2023metric3d,bochkovskii2024depth} aim to recover depth from a single image; however, even state-of-the-art methods still exhibit substantial errors, often ranging from several to tens of meters. In contrast, relative depth estimation in a zero-shot manner provides a viable alternative by predicting normalized disparity $\overline{t}$, which preserves ordinal depth relationships between scene elements (e.g., "object A is closer than object B") and remains robust without requiring metric supervision or dataset-specific fine-tuning. We first describe the transformation relationships among depth, disparity, and normalized disparity here. Let the depth at pixel $(u,v)$ be denoted by $d_{u,v}$. The disparity is defined as its inverse, $t_{u,v} = 1 / d_{u,v}$, and the normalized disparity $\overline{t}_{u,v}$ for each image is given by:
\begin{equation}\label{eq:norm}
\overline{t}_{u,v} = \frac{t_{u,v}-\min(t)}{\max(t)-\min(t)}, \quad \overline{t}_{u,v}\in(0,1),
\end{equation}
where $\max(t)$ and $\min(t)$ denote the extrema of disparity in the image. Intuitively, $\overline{t}_{u,v}$ can be viewed as a probability-like quantity that distributes pixels along a $(0,1)$ interval, thereby capturing the structural layout of the scene without requiring metric consistency across images. \\
With the pseudo ground-truth normalized disparity $\hat{t}_{u,v}$ generated from  DAM decoder~\citep{yang2024depth}, we thus transform the predicted depth $d_{u,v}$ into normalized disparity $\overline{t}_{u,v}$ with Eq.~(\ref{eq:norm}) first, and then supervise the predicted $\overline{t}_{u,v}$ against $\hat{t}_{u,v}$ using an L1 loss on the image:
\begin{equation}
\mathcal{L}_{dis} = \frac{1}{UV}\sum_{u=1}^U\sum_{v=1}^V\big|\overline{t}_{u,v}-\hat{t}_{u,v}\big|,
\end{equation}
which encourages the network to learn the ordinal depth distribution while transferring geometric priors from DAM into our framework. For intuitive understanding, the entire process of DDA is depicted in Fig.~\ref{fig:c2b}.

\subsection{Semantic-guided neural localization}\label{sec:loc}

\textbf{Semantic Alignment:}  Since OSM data provides semantic and geographic information about stationary objects, the key to successful localization lies in constructing the BEV feature $\boldsymbol{F}_{bev}$ with consistent semantics and topology to those of the OSM data. Explicit feature registration at the semantic level also aligns with how humans localize themselves~\citep{epstein2014neural}. However, previous I2O visual localization approaches largely overlooked the semantic consistency of images and OSM data. OrienterNet~\citep{sarlin2023orienternet} exploited the network for pose estimation in an end-to-end way, whereas pose supervision only makes it challenging for the model to fully understand the surroundings. Concurrent approach MaplocNet~\citep{wu2024maplocnet} pointed out that the semantic labels from the HD maps could improve the comprehension ability of the image model and substantially enhance visual localization accuracy. However, these semantic labels are not freely accessible and limit the method's generalizability. \\
To address the challenge, we introduce an auxiliary semantic alignment task to improve the scene understanding capability of the image features. Intuitively, with the ground-truth pose $\hat{\boldsymbol{p}}$, the proposed task crops the corresponding areas from the OSM semantic embeddings $\boldsymbol{\mathcal{M}}_{sem}$ and uses it to guide the BEV image feature $\boldsymbol{F}_{bev}$ learning the semantic distribution. For each pixel $(d_i,l)$ on the BEV feature $\boldsymbol{F}_{bev} \in \mathbb{R}^{D \times L \times C}$, we estimate the corresponding grid $\hat{\Gamma}(d_i,l)$ on the map semantic embeddings $\boldsymbol{\mathcal{M}}_{sem}$ using the transformation matrix $\hat{\Gamma}=\Gamma(\hat{\boldsymbol p}) \in SE(2)$ of the ground-truth pose $\hat{\boldsymbol p} \in \mathbb{R}^3$. We then directly align the BEV features $\boldsymbol{F}_{bev}$ with the corresponding semantic embeddings of the OSM data. The semantic alignment loss $\mathcal{L}_{sem}$ is defined as:
\renewcommand{\theequation}{6}
\begin{equation}
    \mathcal{L}_{sem} = \frac{1}{DL} \sum_{i=1}^D \sum_{l=1}^L  
    \big\| C_{1\times1}(\boldsymbol{F}_{bev}[d_i,l]) - \boldsymbol{M}_{sem}[\hat{\Gamma}(d_i,l)] \big\|_2,
\end{equation}
where $C_{1 \times 1}$ denotes a $1\times1$ convolution layer for channel adjustment. The proposed semantic alignment task aligns BEV features with OSM semantic embeddings, enabling the image network to better perceive its surroundings in the BEV space. This enhanced perception benefits neural matching-based localization and comes essentially free of additional annotation. 

 With the semantic-rich BEV features $F_{bev}$ and map features $F_{map}$, the neural localization is responsible for estimating the 3-DoF pose $\boldsymbol{p}=(x,y,\theta)$ of the image. The most straightforward strategy is, fusing the BEV features and map features and estimating the pose with a small regressor. Unfortunately, in our pilot studies, the network failed to converge with the regression design after trial and error, which contradicts the observation in MaplocNet~\citep{wu2024maplocnet}. We speculate that this is due to the limited field of view (FoV) of a single image, which results in the BEV feature $\boldsymbol{F}_{bev}$ containing partial information. Therefore, directly learning the pose vector is challenging. To simplify the problem, we discretely sample the continuous 3-D pose space at each map grid and $K$ rotations with regular intervals, resulting in a $H\times W \times K$ volume of pose candidates in the 3-DoF pose space. For each pose candidate $\boldsymbol{p}_{h,w,k}=(x_h,y_w,\theta_k)$, we project the BEV feature $\boldsymbol{F}_{bev}$ onto the map plane and calculate the matching score in the feature space :
 \begin{equation}\label{eq:fea_match}
     \boldsymbol{S}_{h,w,k} = \frac{1}{DL} \sum_{i=1}^D \sum_{l=1}^L \boldsymbol{F}_{map}(\Gamma_{h,w,k}(d_i,l)) \odot \boldsymbol{F}_{bev}(d_i,l),
 \end{equation}
 where $\Gamma_{h,w,k}$ is the transformation matrix for pose candidate $\boldsymbol{p}_{h,w,k}$, and $\odot$ denotes the inner product operation.
 Eq.~(\ref{eq:fea_match}) benefits from an efficient implementation by performing a single convolution as a batched multiplication in the Fourier domain~\citep{barsan2020learning}. 
Finally, the scoring volume $\boldsymbol{S}$ is normalized to the probability volume $\boldsymbol{P}$ with the softmax operation.

The pose loss $\mathcal{L}_{pose}$ is defined as a binary cross-entropy loss to maximize the probability of the GT pose $\hat{\boldsymbol{p}}$:
\begin{equation}
    \mathcal{L}_{pose} = - \sum_{h=1}^H \sum_{w=1}^W \sum_{k=1}^K \hat{P}_{h,w,k} \log(P_{h,w,k}) = - \mathrm{log} P_{\hat{x},\hat{y},\hat{\theta}},
\end{equation}
where the $\hat{P}$ is the probability volume of the GT pose, which is concentrated at a single point $\boldsymbol{\hat{p}}=(\hat{x},\hat{y},\hat{\theta})$ in the pose space.

Overall, the joint loss $\mathcal{L}$ is defined as a weighted combination of  pose loss $\mathcal{L}_{pose}$, disparity loss $\mathcal{L}_{dis}$ and semantic alignment loss $\mathcal{L}_{sem}$:
\begin{equation}
    \mathcal{L} = \lambda_1 \mathcal{L}_{pose} + \lambda_2 \mathcal{L}_{dis} + \lambda_3 \mathcal{L}_{sem},
\end{equation}
where the $\lambda_1$, $\lambda_2$, $\lambda_3$ are hyper-parameters to control the weights of each component.

During inference, we estimate the pose $\boldsymbol{p}^*$ as the one corresponding to the maximal value in the probability volume $\boldsymbol{P}$:
\begin{equation}\label{eq:infer}
   \boldsymbol{p}^*=\mathrm{argmax}_{\boldsymbol{p}} \, P(\boldsymbol{p}|\mathcal{I},\mathcal{M}) 
\end{equation}

\section{Cross-area and cross-condition validation benchmark}\label{sec: CC}
OrienterNet~\citep{sarlin2023orienternet} has constructed the MGL dataset for the I2O visual localization task, containing 826\,K images for training and 2\,K images for validation. The training-evaluation split follows the "same-area" protocol, where each location's images are disjointed into training and validation sets. Besides, the distribution of cameras, motions, viewing conditions, and visual features of the validation set is identical to the training set. 
Such \textbf{same-area} and \textbf{same-condition} setup may work well for localization within a small, local area over a short time span, it is not appropriate for assessing the performance of I2O localization methods across different countries with VGI data collected at various conditions.

\begin{figure*}[!ht]
\centering
\includegraphics[width=6.8in]{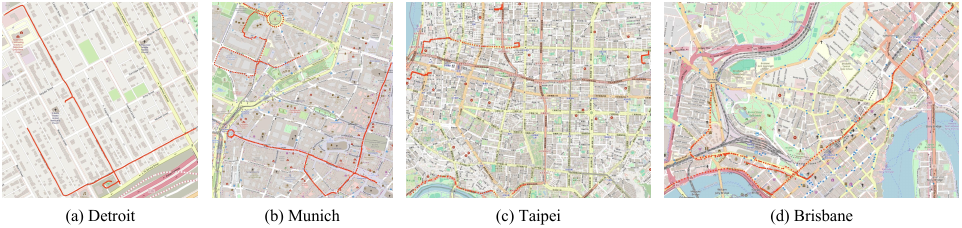}
\caption{OSM data coverage and image trajectories in the CC validation benchmark. The red points represent the locations of the images.}
\label{fig:benchmark}
\end{figure*}

\begin{figure}[h]
\centering
\includegraphics[width=3.2in]{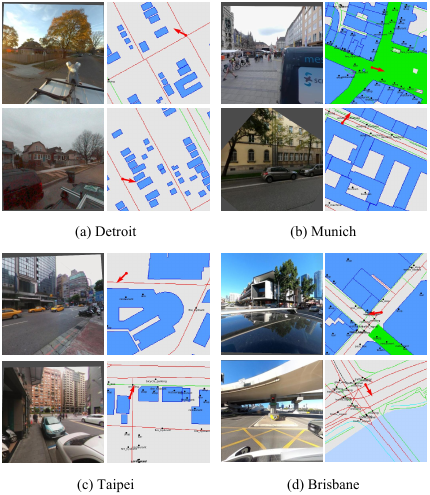}
\caption{Query image and map tile pairs from the CC validation benchmark. Red arrow \( \boldsymbol{\textcolor{red}{\raisebox{-0.5ex}{\rotatebox{90}{$\rightarrow$}}}} \) represents the GT pose in the map tile.}
\label{fig:pair_cc}
\end{figure}

To address this concern, we propose a novel validation benchmark for \textbf{cross-area} and \textbf{cross-condition} validation, called the \textbf{CC} validation benchmark in the following. Table. \ref{table: benchmark} lists details of the CC validation benchmark, which contains 7164 images from 4 cities on the Mapillary platform~\citep{Mapillary_website}. These images were collected with diverse devices, by different users, with varying motion patterns and under various external conditions, as shown in Fig.~\ref{fig:benchmark} and Fig.~\ref{fig:pair_cc}. They have never been seen before and are located hundreds of miles away from the nearest area in the MGL dataset. Each image is captured by cameras handheld or mounted on helmets, UGVs, or vehicles. For each 360$^\circ$ panorama, we split it into $4 \times 90^\circ$ FoV perspective images with a random offset and resample them to the size of $512 \times 512$ pixels. We query the OSM data for the corresponding area from the official website~\citep{OSM_website} and the data pre-processing is shown in Fig.~\ref{fig:rasterization}. The validation scenes in the CC validation benchmark cover a broad range of urban scenes, including downtown, suburban, overpass regions, and pedestrian streets.  We believe the diverse set of scenes, captured by various cameras, users, platforms, and under different external conditions, makes the CC benchmark more challenging and suitable for comprehensive evaluation.

\begin{table}[ht]
\caption{Details of CC validation benchmark.}
\label{table: benchmark}
\footnotesize
\setlength{\tabcolsep}{9.0pt}
\begin{tabular}{cccc}
\hline
City                 & Image               & Camera            & Platform               \\ \hline
\multirow{2}{*}{Detroit}        & \multirow{2}{*}{1676} & Trimble mx50        & \multirow{2}{*}{Vehicle} \\
                                    &                       & Gray Search Ladybug &                          \\ \hline
\multirow{2}{*}{Munich}     & \multirow{2}{*}{2156} & Trimble mx50        & Helmet                 \\
                                    &                       & Labpano Pilot Era   & UGV                      \\ \hline
\multirow{2}{*}{Taipei}         & \multirow{2}{*}{1872} & LG-R105             & Handheld                 \\
                                    &                       & RICOH THETA V       & Vehicle                  \\ \hline
\multirow{2}{*}{Brisbane} & \multirow{2}{*}{1460} & Garmin VIRB 360     & \multirow{2}{*}{Vehicle} \\
                                    &                       & GoPro MAX           &                          \\ \hline
\end{tabular}
\end{table}

\section{Experiment}
\subsection{Experimental setup}
 We evaluate the performance of our OSMLoc on three datasets: the  MGL dataset, the CC validation benchmark, and the KITTI~\citep{geiger2012we} dataset. Our model is trained on the MGL dataset and directly validated on the CC validation benchmark and KITTI dataset without fine-tuning.

\begin{itemize}
    \item MGL dataset: consists of 2580 sequences with approximately 828\,K images collected in 13 cities from the Mapillary platform, which exposes the camera calibration, noise GPS measurement, and the 3-DoF pose in a global reference frame obtained by the fusion of motion pattern and GPS. All the sequences were recorded after 2017 with cameras known for high-quality reconstructions. Images of each city are split into training and validation sets, resulting in 826\,K training and 2\,K validation images. 
    
    \item  CC benchmark: We construct the \textbf{cross-area} and \textbf{cross-condition} (CC) benchmark with 7,164 images from four unseen cities: Detroit, Munich, Taipei, and Brisbane, which are located hundreds of kilometers from the MGL training scenes. This benchmark is designed to assess the global generalization capability of I2O approaches.
    
    \item KITTI dataset: KITTI~\cite{geiger2012we} provides stereo images captured by a moving vehicle in Karlsruhe, Germany, which is utilized to demonstrate the application value of our OSMLoc in the autonomous driving field and various image-capturing platforms. \citet{shi2022beyond} split the KITTI dataset into the \textit{Train}, \textit{Test1} and \textit{Test2} sets for I2A visual localization task. The \textit{Test2} set was collected in a different area, with no overlap of the MGL dataset or the KITTI \textit{Train} set. Following the OrienterNet~\citep{sarlin2023orienternet}, we select the \textit{Test2} as the validation set, assuming that an accurate initial pose sampled within $\pm 20\text{m}$ and $\pm 10^\circ$  is available. Only the pose candidates that fall within the range are considered during inference. 
\end{itemize}

\textbf{Baseline methods:} Due to the limited focus on the single-image I2O visual localization, OrieneterNet~\citep{sarlin2023orienternet} is currently the only open-sourced comparable method in the field. We adopt variants of I2A visual localization methods~\citep{shi2022beyond,xia2022visual,sarlin2024snap}, along with the foundational model based method AnyLoc~\citep{keetha2023anyloc}, as baselines for comparison.
\begin{itemize}
    \item OrienterNet~\citep{sarlin2023orienternet}: is the pioneering end-to-end I2O visual localization method for single image. The image encoder is a ResNet101-FPN~\citep{he2016deep} pre-trained on ImageNet, while the map encoder is a VGG19-FPN~\citep{simonyan2014very} architecture trained from scratch.  We use the officially released code for implementation and retrain the model with the same settings, except for setting the batch size to 4.
    \item Retrieval~\citep{xia2022visual}: replaces BEV inference and matching with a correlation between the neural map and a global image embedding. While the original implementation only estimates a 2-DoF position without orientation, we extend it by predicting orientation through four neural maps corresponding to the \textit{North}, \textit{South}, \textit{East}, and \textit{West} directions.
    \item Refinement~\citep{shi2022beyond}: updates an initial pose by warping a satellite view to the image, assuming that the scene is planar, at a fixed height, and gravity-aligned. We replace the satellite image with a rasterized OSM tile. This method requires an initial orientation during both training and evaluation, which we sample within $45^\circ$ of the ground truth orientation angle.
    \item SNAP~\citep{sarlin2024snap}: is a ground image-to-aerial image localization method. Following the official implementation, we utilize a ResNet50-FPN network as the image encoder and a VGG19-FPN network as the map encoder. InfoNCE~\citep{oord2018representation} loss is introduced to maximize the correct localization probability and minimize the mismatch probability.
    \item AnyLoc~\citep{keetha2023anyloc}: is originally developed for visual place recognition and leverages the foundational model DINOv2~\cite{oquab2023dinov2} to avoid re-training or fine-tuning. However, since OSM tiles contain only semantic information rather than color or texture cues, we observed that the foundation model fails to extract sufficiently discriminative features. To address this, we employ the DINOv2-S model as the image encoder and NetVLAD as the feature aggregator for end-to-end training, while adopting a VGG19-FPN architecture as the map encoder. To handle orientation, we further predict rotation by generating four neural maps corresponding to the \textit{North-South-East-West} directions.  
\end{itemize}

\textbf{Evaluation metrics:} For the MGL dataset and the CC validation benchmark, we calculate the position recall (PR), orientation recall (OR), absolute position error (APE) and absolute orientation error (AOE) as the evaluation metrics. With the predicted pose $\boldsymbol{p}_i=( x_i, y_i,  \theta_i)$ and GT pose $\boldsymbol{\hat{p}}_i=( \hat{x}_i, \hat{y}_i,  \hat{\theta}_i)$ of frame $i$, the PR and OR are formulated as :
\begin{equation}
    \begin{aligned}
    & \text{PR}=\frac{1}{N} \sum_{i=1}^N{\mathds{1} (\| (x_i, y_i) - (\hat{x}_i,\hat{y}_i) \|_2 < \sigma_p),} \\
    & \text{OR}=\frac{1}{N} \sum_{i=1}^N \mathds{1} (| \theta - \hat{\theta}_i | < \sigma_o),
\end{aligned}
\end{equation}

where $N$ is the number of test frames and the threshold pairs $\boldsymbol{\sigma}=(\sigma_p,\sigma_o)$ are the maximal tolerance of the localization error.  The APE  and AOE are the mean absolute error between the predicted poses and GT poses:
\begin{equation}
\begin{aligned}
 & \text{APE} = \frac{1}{N}\sum_{i=1}^{N} \| (x_i,y_i) - (\hat{x}_i,\hat{y}_i) \|_2, \\
 & \text{AOE} = \frac{1}{N}\sum_{i=1}^{N} |  \theta_i - \hat{\theta}_i |. 
\end{aligned}
\end{equation}
For the KITTI dataset, since position recall is closely related to the vehicle's motion direction in autonomous driving scenes~\citep{shi2022beyond}, we decompose the position recall and error into two components relative to the viewing axis: recall and absolute error in the perpendicular (latitudinal) direction (LatR and ALatE) and in the parallel (longitudinal) direction (LonR and ALonE).

\textbf{Implementation details:} 
  For the image, we set $U$ and $V$ to half of the original image size and  $D=64$, $L=129$,  $d_0=0$, $\Delta=0.5\text{m}$ during the camera-to-BEV transformation. For the OSM data, we render map tiles of size $H \times W=128\text{m} \times 128 \text{m}$ centered around points randomly sampled within 32\text{m} of the ground truth pose.   In the PyTorch implementation, we employ the \texttt{nn.Embedding} layer to map each channel to a $C_{sem}=16$-dimensional feature vector.  The orientation sampling number $K$ is set to 64 during training and 256 during inference.   The GT pose $\boldsymbol{\hat{p}}$, pseudo GT normalized disparity $\boldsymbol{\hat{t}}$ and the semantic embedding $\boldsymbol{\mathcal{M}}_{sem}$ are jointly used to supervise the network.  We trained the network for 500K steps with a batch size of 4, using SGD to optimize the network, and the initial learning rate is set to 1e-4. The hyperparameters for the loss weight are set to $\lambda_1=1, \lambda_2=20, \lambda_3=10$. The threshold pairs $\boldsymbol{\sigma}=(\sigma_p,\sigma_o)$ are set to $(1\text{m}, 1^\circ)$, $(3\text{m}, 3^\circ)$ and $(5\text{m}, 5^\circ)$ during evaluation.   All the experiments are conducted on a desktop with an Intel i9-13900K CPU and an NVIDIA GeForce RTX 4090 GPU. By incorporating the small and base variants of DINOv2 as the image encoder and adjusting the channels and layers of the FPN decoder, we design two variants of our method: the small and base versions (denoted as OSMLoc-S and OSMLoc-B, respectively).  More configurations and implementation details are available in our open-source code.

\subsection{Localization results}

\begin{table*}[!t]
\footnotesize
\setlength{\tabcolsep}{10.0pt}
\caption{Visual localization results on the MGL dataset. $\uparrow$ means higher is better, $\downarrow$ means lower is better. The localization results of "Retrieval"~\citep{xia2022visual} and "Refinement"~\citep{shi2022beyond} are taken from OrienterNet~\citep{sarlin2023orienternet}. We use \textbf{bold} font to indicate the best results and \underline{underline} to indicate the second-best results.}
\label{table: MGL_results}
\begin{tabular}{ccccccccccc}
\hline
\multirow{2}{*}{Method}  & \multicolumn{3}{c}{PR @Xm$\uparrow$}                & \multicolumn{3}{c}{OR @X$^\circ$$\uparrow$}      & \multirow{2}{*}{APE($\text{m}$)$\downarrow$} & \multirow{2}{*}{AOE($^\circ$)$\downarrow$}  &  \multirow{2}{*}{FPS$\uparrow$}   \\ \cline{2-7} 
                        & @1m            & @3m            & @5m            & @1$^\circ$             & @3$^\circ$             & @5$^\circ$            \\ \hline
Retrieval                & 2.02           & 15.21          & 24.21          & 4.50           & 18.61          & 32.48     & - & -  & -   \\
Refinement             & 8.09           & 26.02          & 35.31          & 14.92          & 36.87          & 45.19      & - & -  & - \\
OrienterNet           & 10.69          & 42.12          & 54.01          & 19.22          & 48.56          & 63.59      & 12.00 & 28.71  & 10.77\\
 AnyLoc    &  0.06           &  0.31          &  0.78           &  0.52           &  1.78           &  2.93       &  55.32  &  87.78   &  \textbf{13.10}  \\
 SNAP    &  9.11           &  37.51           &  49.08           &  19.49           &  48.04           &  62.23       &  13.20  &  31.09   & \underline{12.18}  \\
OSMLoc-S          & \underline{14.04} & \underline{44.00} & \underline{57.31} & \textbf{21.84} & \underline{53.38} & \underline{68.41} & \underline{11.09} & \underline{26.09} & 11.13 \\
OSMLoc-B           & \textbf{15.30} & \textbf{46.05} & \textbf{59.40} & \underline{21.79} & \textbf{56.57} & \textbf{71.29} & \textbf{10.04} & \textbf{22.91} & 9.27 \\\hline
\end{tabular}
\end{table*}

\begin{figure}[ht]
\centering
\includegraphics[width=3.3in]{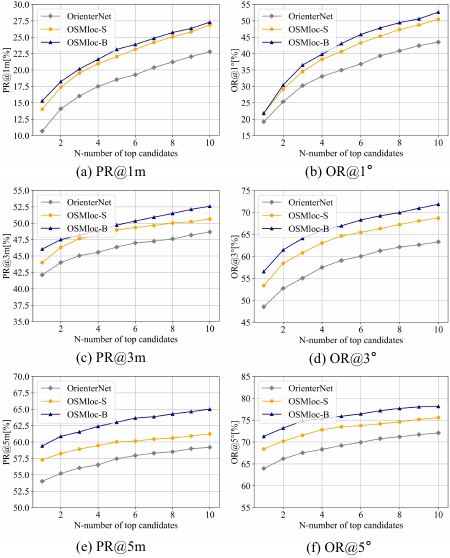}
\caption{Position and orientation recall curves of the top candidates on the MGL dataset. The gray, orange, and navy curves are for OrienterNet, OSMLoc-S and OSMLoc-B respectively. }
\label{fig:PR_OR_curve}
\end{figure}

\textbf{MGL dataset:} Table.~\ref{table: MGL_results} shows the single-image visual localization results on the MGL dataset. Notably, our OSMLoc significantly outperforms all baseline methods by a large margin across all position and orientation recall thresholds. Specifically, OSMLoc-S reduces the APE and the AOE by approximately $0.91\text{m}/2.62^\circ$ and improves the $5\text{m}/5^\circ$ recall by $3.30\%/4.82\%$ over the SOTA baseline method, OrienterNet~\citep{sarlin2023orienternet}; OSMLoc-B achieves the best performance, with a  $1.96\text{m}/5.80^\circ$ reduction in APE and AOE and a $5.39\%/7.70\%$ improvement in the $5\text{m}/5^\circ$ recall. With comparable inference speed, the improvement mainly results from incorporating geometric and semantic guidance into the I2O localization framework rather than enlarging the model. Furthermore, Fig.~\ref{fig:PR_OR_curve} shows the recall curves for top candidates at different thresholds, where OSMLoc-B achieves the highest recall across various settings. These quantitative results demonstrate that our method significantly enhances prediction accuracy and robustness, as evidenced by consistent gains in position and orientation recall across different thresholds and top candidates.

\begin{figure*}[!t]
\centering
\includegraphics[width=6.8in]{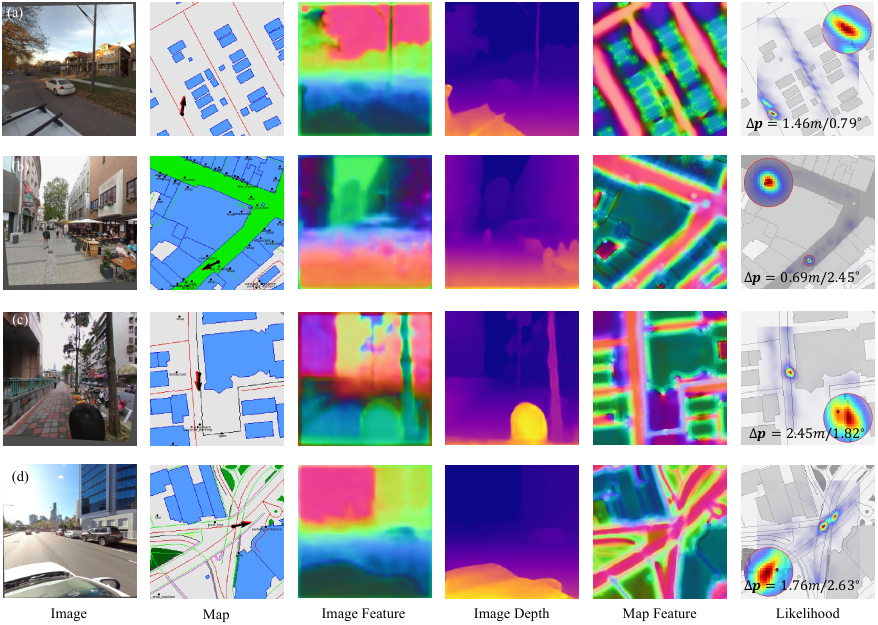}
\caption{Qualitative results of OSMLoc-S on the CC validation benchmark. (a) Detroit. (b) Munich. (c) Taipei. (d) Brisbane. Black arrow 
\( \boldsymbol{\textcolor{black}{\raisebox{-0.5ex}{\rotatebox{90}{$\rightarrow$}}}} \)
 represents the predicted pose and red arrow \( \boldsymbol{\textcolor{red}{\raisebox{-0.5ex}{\rotatebox{90}{$\rightarrow$}}}} \) represents the GT pose in the map.}
\label{fig:qua_ca_mgl}
\end{figure*}

\textbf{CC validation benchmark:}
We evaluate our method and the baseline method OrienterNet, on the CC validation benchmark and report the single-image localization results in Table.~\ref{table: cross_area}. Although the CC validation benchmark is more challenging than the same-area and same-condition validation set of the MGL dataset, our method consistently outperforms the baseline across all four cities, demonstrating a clear advantage in various scenes. Specifically, our OSMLoc-B achieves $16.77\%/17.84\%$ and $5.47\%/12.25\%$ improvement of $5\text{m}/5^\circ$ recall in Detroit and Munich respectively, demonstrating the superiority of our method. Since the test areas of Taipei and Brisbane are situated in central urban districts with numerous skyscrapers and overpasses, these scenes differ significantly from the training set and offer limited useful information for localization. We believe that these factors contribute to the relatively lower localization success rate in these regions. Nevertheless, our OSMLoc still achieves significantly higher localization accuracy compared to the baseline. Fig.~\ref{fig:qua_ca_mgl} presents the qualitative results of OSMLoc-S on the CC validation benchmark. 

\begin{table}[ht]
\footnotesize
\setlength{\tabcolsep}{2.5pt} 
\caption{Visual localization results on the CC validation benchmark. $\uparrow$ means higher is better. We use \textbf{bold} font to indicate the best results.}
\label{table: cross_area}
\begin{tabular}{cccccccc}
\hline
\multirow{2}{*}{City}    & \multirow{2}{*}{Method} & \multicolumn{3}{c}{PR @Xm$\uparrow$}                       & \multicolumn{3}{c}{OR @X$^\circ$$\uparrow$}                        \\ \cline{3-8} 
                         &                           & @1m            & @3m            & @5m            & @1$^\circ$             & @3$^\circ$             & @5$^\circ$             \\ \hline
\multirow{3}{*}{Detroit}   & OrienterNet               & 7.64           & 33.11          & 46.24          & 13.66          & 32.10          & 44.57          \\
                   &       OSMLoc-S             & 11.99 & 43.32 & 52.45 & 16.77 & 39.86 & 54.47 \\ 
                   &       OSMLoc-B             & \textbf{13.37} & \textbf{51.79} & \textbf{63.01} & \textbf{19.75} & \textbf{48.57} & \textbf{62.41} \\ \hline
\multirow{3}{*}{Munich}   & OrienterNet               & 2.68           & 17.90          & 35.11          & 12.48          & 32.75           & 47.40          \\
                        &   OSMLoc-S             & 3.20  & 19.53 & 38.27 & 15.07 & 39.19  & 55.94 \\
                        &   OSMLoc-B             & \textbf{4.36}  & \textbf{22.03} & \textbf{40.58} & \textbf{15.35} & \textbf{41.79}  & \textbf{59.65} \\ \hline
\multirow{3}{*}{Taipei}    & OrienterNet               & 1.23           & 8.17           & 16.13          & 5.88           & 16.40         & 24.79          \\
                          &  OSMLoc-S              & 1.07  & 8.39 & 18.00 & 7.00  & 20.35 & 29.59 \\
                          &  OSMLoc-B              & \textbf{1.44}  & \textbf{9.72} & \textbf{19.18} & \textbf{7.32}  & \textbf{21.31} & \textbf{32.10} \\
                          \hline

\multirow{3}{*}{Brisbane} & OrienterNet               & 0.21           & 2.60           & 7.47           & 4.66           & 9.73          & 17.60          \\
                          &   OSMLoc-S              & \textbf{0.41}  & 2.81  & 8.56  & \textbf{5.55}  & 15.34 & 23.90 \\
                          &   OSMLoc-B              & 0.27  & \textbf{2.88}  & \textbf{9.18}  & 5.00  & \textbf{15.34} & \textbf{24.86} \\ 
                          \hline
\end{tabular}
\end{table}

\begin{figure*}[h]
\centering
\includegraphics[width=6.6in]{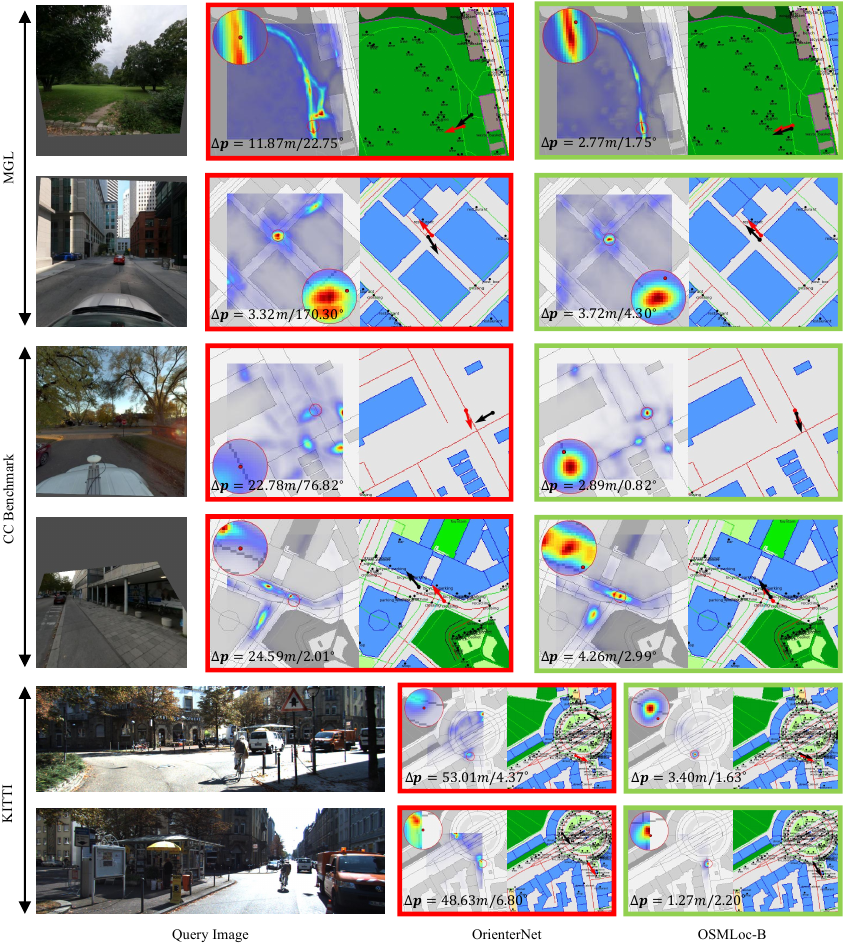}
\caption{Qualitative results on the MGL dataset, CC validation benchmark and KITTI dataset. The red rectangle \( \raisebox{0.5ex}{\textcolor{red}{\fbox{}}} \) represents the wrong localization result and the green rectangle represents \( \raisebox{0.5ex}{\textcolor{green}{\fbox{}}} \) the correct localization result. Black arrow
\( \boldsymbol{\textcolor{black}{\raisebox{-0.5ex}{\rotatebox{90}{$\rightarrow$}}}} \)
 represents the predicted pose and red arrow \( \boldsymbol{\textcolor{red}{\raisebox{-0.5ex}{\rotatebox{90}{$\rightarrow$}}}} \) represents the GT pose in the map.}
\label{fig:qua_compare_all}
\end{figure*}

\textbf{KITTI dataset:} Extensive experiments are conducted on the KITTI dataset to demonstrate the generalizability of our method in self-driving scenarios. The quantitative results are shown in Table.~\ref{tab:kitti}. Without fine-tuning, our OSMLoc outperforms baseline methods trained on the KITTI dataset by a large margin, demonstrating its strong generalization capability and robustness. Furthermore, compared with the methods~\citep{shi2022beyond,xia2022visual,sarlin2024snap,keetha2023anyloc}
trained on the MGL dataset, our OSMLoc achieves much higher position and orientation recall. These quantitative results highlight that our method excels in accuracy, robustness, and generalization, making it suitable for versatile downstream tasks. Moreover, our OSMLoc-S reduces the absolute latitudinal error (ALatE), longitudinal error (ALonE) and orientation error (AOE)  from 1.35\text{m}/6.64\text{m}/2.50$^\circ$ to 1.28\text{m}/5.70\text{m}/1.99$^\circ$, respectively. Meanwhile, the OSMLoc-B achieves the lowest localization error at 1.15\text{m}/5.42\text{m}/1.99$^\circ$.  Fig.~\ref{fig:qua_compare_all} deploys additional qualitative results, where our method produces cleaner likelihood maps and more accurate camera pose predictions.

\begin{table*}[ht]
\caption{Quantitative results on the KITTI dataset. We use \textbf{bold}
font to indicate the best results and \underline{underline} to indicate the second-best.}
\footnotesize
\setlength{\tabcolsep}{6.0pt}
\label{tab:kitti}
\begin{tabular}{cccccccccccc}
\hline
\multirow{2}{*}{Map}       & \multirow{2}{*}{Method} & \multirow{2}{*}{Training dataset}  & \multicolumn{3}{c}{LatR @Xm$\uparrow$}                                                                                      & \multicolumn{3}{c}{LonR @Xm$\uparrow$}                                                                                   & \multicolumn{3}{c}{OR @X$^\circ$$\uparrow$}                                                                                       \\ \cline{4-12} 
                           &                                                                                                      &                                                                          & @1m                                & @3m                                & @5m                                & @1m                                & @3m                                & @5m                                & @1$^\circ$                                 & @3$^\circ$                                 & @5$^\circ$                                 \\ \hline
\multirow{3}{*}{Satellite} & DSM                     & \multirow{3}{*}{KITTI}                                                                                                                                 & 10.77                              & 31.37                              & 48.24                              & 3.87                               & 11.73                              & 19.50                              & 3.53                               & 14.09                              & 23.95                              \\
                           & VIGOR                   &                                                                                                                                                     & 17.38                              & 48.20                              & 70.79                              & 4.07                               & 12.52                              & 20.14                              & -                                  & -                                  & -                                  \\
                           & Refinement              &                                                                                                                                                       & 27.82                              & 59.79                              & 72.89                              & 5.75                               & 16.36                              & 26.48                              & 18.42                              & 49.72                              & 71.00                              \\ \hline
\multirow{5}{*}{OSM}       & Retrieval               & \multirow{5}{*}{MGL}                                                                                                                                 & 37.47                              & 66.24                              & 72.89                              & 5.94                               & 16.88                              & 26.97                              & 2.97                               & 12.32                              & 23.27                              \\
                           & Refinement              &                                                                                                                                                       & 50.83                              & 78.10                              & 82.22                              & 17.75                              & 40.32                              & 52.40                              & 31.03                              & 66.76                              & 76.07                              \\
                           & OrienterNet             &                                                                                                                                                       & 62.03          & 90.89          & 95.73          & 19.99          & 51.60          & 65.78          & 29.93          & 72.81          & 87.79          \\ 
                           &  AnyLoc              &                                                                                                                                                       &  3.12           &  9.57           &  15.73           &  1.38          &  4.49           &  7.78          &  5.46          & 14.76          &  24.64           \\ 
                           &  SNAP             &                                                                                                                                                       & 41.05          & 84.02          & 92.64          & 22.86         & 53.53          & 64.56         & 25.32          & 60.63          & 79.67          \\
                           & OSMLoc-S           &                                                                                                                                                        & \underline{65.75} & \underline{93.79} & \underline{96.74} & \underline{24.50} & \underline{59.53} & \underline{72.67} &  \textbf{37.43} & \textbf{79.69} & \underline{91.67}  \\
                           & OSMLoc-B  &                                                                           & \textbf{66.60} & \textbf{95.05} & \textbf{97.41} & \textbf{29.40} & \textbf{63.86} & \textbf{74.52} &  \underline{35.57} & \underline{78.28} & \textbf{92.96} \\ 
                           \hline
\end{tabular}
\end{table*}

\begin{table}[ht]
\footnotesize
\setlength{\tabcolsep}{2.0pt}
\caption{Ablation studies on the foundation model of image encoder. The "FLOPs" metric includes only the computational complexity of the image encoder for fair comparison. ResNet101-FPN*: Image encoder employed by OrienterNet.}
\label{table: foundation_model}
\begin{tabular}{cccccccc}
\hline
 \multirow{2}{*}{Image Encoder} & \multicolumn{3}{c}{PR @Xm$\uparrow$} & \multicolumn{3}{c}{OR @X$^\circ\uparrow$} & \multirow{2}{*}{FLOPs$\downarrow$} \\ \cline{2-7}
                               & @1m    & @3m   & @5m   & @1$^\circ$     & @3$^\circ$     & @5$^\circ$         &                     \\ \hline
ResNet-101   &  12.90                &   43.27     &  56.05     &  20.48     & 52.28       &     66.42                          &   52.21G  \\
DINOv2-S   & 14.04                   &   44.00     &  57.31     &  $\boldsymbol{21.84}$      &  53.38     & 68.41        &       $\boldsymbol{48.16}$G               \\
DINOv2-B   &  $\boldsymbol{15.30}$       &  $\boldsymbol{46.05}$     & $\boldsymbol{59.40}$      & 21.79       &  $\boldsymbol{56.57}$     & $\boldsymbol{71.29}$  &    154.10G                          \\ \hline
\end{tabular}
\end{table}

\begin{table}[H]
\caption{Ablation studies on geometric guidance and semantic guidance. OSMLoc-S is selected as the baseline. "Geo" refers to the geometric guidance for camera-to-BEV transformation, while "Sem" denotes the semantic guidance from the OSM embeddings for cross-modal alignment.}
\label{table: guidance}
\footnotesize
\centering
\setlength{\tabcolsep}{3pt}
\begin{tabular}{ccccccccccc}
\hline
\multirow{2}{*}{}      &  \multirow{2}{*}{Geo} & \multirow{2}{*}{Sem} & \multicolumn{3}{c}{PR @Xm$\uparrow$}                & \multicolumn{3}{c}{OR @X$^\circ\uparrow$ }             \\ \cline{4-9} 
        &   &                                    & @1m            & @3m            & @5m            & @1$^\circ$             & @3$^\circ$             & @5$^\circ$             \\ \hline
$[A]$                                     &                &                &     11.52           &    42.54            &   55.53     & 20.80  & 51.44 &  65.79      \\ 
$[B]$                                     &      \checkmark           &                &     12.33           &    43.22            &   55.15     & 21.22  & 51.74 &  66.01         \\
$[C]$                                      &                 &   \checkmark             &     13.20           &    43.43            &   56.16     & 20.43  & 51.08 &  66.16         \\
$[D]$              & \checkmark    &   \checkmark                         & \textbf{14.04} & \textbf{44.00} & \textbf{57.31} & \textbf{21.84} & \textbf{53.38} & \textbf{68.41} \\ \hline
\end{tabular}
\end{table}

\begin{table}[ht]
\footnotesize
\caption{Computational complexity, parameters and inference speed.}
\label{tab:computational complexity}
\begin{tabular}{cccc}
\hline
Method      & \#Params & FLOPS   & FPS   \\ \hline
OrienterNet & 54.94M   & 89.63G  & 10.77      \\
SNAP        & 35.95M   & 70.17G  & 12.18 \\
AnyLoc      & 35.38M   & 71.21G  & 13.10  \\
OSMLoc-S    & 37.57M   & 85.57G  & 11.13 \\
OSMLoc-B    & 110.60M  & 190.68G & 9.27  \\ \hline
\end{tabular}
\vspace{-12pt}
\end{table}

\subsection{Ablation studies and Analysis}
In this section, we analyze the impact of the foundation model in the image encoder, as well as the geometric and semantic guidance, on the overall framework. Ablation experiments are conducted on the MGL dataset to assess the effectiveness of each module. We also present some failure cases to explore the limitations of our approach. 

\textbf{Impact of the foundation model:} Firstly, we conduct ablation studies to investigate the influence of the foundational model-based image encoder, as shown in Table.~\ref{table: foundation_model}. For the ResNet101-FPN as in OrienterNet~\cite{sarlin2023orienternet}, we pre-build the pseudo normalized disparity ground-truth $\hat{\boldsymbol{t}}$ with the DAM-B model and utilize it as geometric guidance. Semantic guidance is obtained from OSM semantic embeddings, as in our approach. With the geometric and semantic guidance, the ResNet101-FPN based variant achieves better localization accuracy than the naive one in the OrienterNet~\citep{sarlin2023orienternet}, which demonstrates the application value of geometric and semantic guidance on general I2O localization frameworks. Furthermore, replacing the original ResNet101-FPN model with the DINOv2-S encoder boosts the localization accuracy with even less computational complexity, which deploys the powerful prior knowledge from foundational models to the I2O localization task. Utilizing the larger DINOv2-B leads to continuous improvement in localization accuracy, but it also increases the computational complexity and reduces the efficiency. 

\textbf{Geometric and semantic guidance:} Table.~\ref{table: guidance} shows the ablation study results on the depth guidance and semantic guidance. In variant $[A]$, the framework learns to localize directly from the pose label $\hat{\boldsymbol{p}}$, without additional guidance. Variant $[B]$ utilizes the geometric guidance from the DAM and introduces the DDA for geometric-guided camera-to-BEV transformation. The localization results show that geometric guidance is beneficial for accurate localization. Variant $[C]$ aligns the BEV features with the corresponding semantic embeddings $\boldsymbol{\mathcal{M}}_{sem}$ from the OSM data pixel-by-pixel, which also boosts the PR performance. Finally, variant $[D]$ demonstrates that fusing both the geometric guidance and the semantic guidance achieves the best localization performance.

\textbf{Impact of loss weights:} We investigate the effect of the hyperparameters $\lambda_1, \lambda_2, \lambda_3$, which weight the pose loss $\mathcal{L}_{pose}$, disparity loss $\mathcal{L}_{dis}$, and semantic loss $\mathcal{L}_{sem}$, by experimenting with different configurations, as reported in Table.~\ref{table: weight}. Variant $[A]$ shows that the vanilla framework with $\mathcal{L}_{pose}$ only, while variants $[B]$ and $[C]$ with disparity loss $\mathcal{L}_{dis}$ and semantic loss $\mathcal{L}_{sem}$ achieve consistent improvement on localization accuracy. As shown in variants $[D], [E], [F]$,  joint loss of pose, disparity and semantic loss boosts the performance, and set $(\lambda_1,\lambda_2,\lambda_3)=(1,20,10)$ achieves the best result.

\begin{table*}[t]
 \caption{Ablation studies on the loss weights. OSMLoc-S serves as the baseline, where $\lambda_1$, $\lambda_2$, and $\lambda_3$ denote the weights for the pose loss $\mathcal{L}_{pose}$, disparity loss $\mathcal{L}_{dis}$, and semantic loss $\mathcal{L}_{sem}$, respectively.}
\label{table: weight}
\footnotesize
\setlength{\tabcolsep}{10.0pt}
\begin{tabular}{cccccccccc}
\hline
\multirow{2}{*}{}      &  \multirow{2}{*}{$\lambda_1$ ($\mathcal{L}_{pose}$)} & \multirow{2}{*}{$\lambda_2$ ($\mathcal{L}_{disparity}$)} & \multirow{2}{*}{$\lambda_3$ ($\mathcal{L}_{sem}$)} &\multicolumn{3}{c}{PR @Xm$\uparrow$}                & \multicolumn{3}{c}{OR @X$^\circ\uparrow$ }             \\ \cline{5-10} 
        &   &      &                              & @1m            & @3m            & @5m            & @1$^\circ$             & @3$^\circ$             & @5$^\circ$             \\ \hline
$[A]$                                     &     1.0           &  0.0   &  0.0            &     11.52           &    42.54            &   55.53     & 20.80  & 51.44 &  65.79      \\ 
$[B]$                                     &      1.0           & 1.0    &  0.0         &     12.33           &    43.22            &   55.15     & 21.22  & 51.74 &  66.01         \\
$[C]$                                      &    1.0             &   0.0     &  1.0      &     13.20           &    43.43            &   56.16     & 20.43  & 51.08 &  66.16         \\
$[D]$              & 1.0  &   1.0      &    1.0              & 12.36 & 43.95 & 56.94 & 20.85 & 51.13 & 66.68 \\ 
$[E]$  & 10.0  &   10.0      &    10.0              & 12.89  & \textbf{44.05}  & 55.94  & 21.00  & \textbf{53.87}  & 67.84 \\ \hline 
$[F]$              & 1.0  &   20.0      &    10.0              & \textbf{14.04} & 44.00 & \textbf{57.31} & \textbf{21.84} & 53.38 & \textbf{68.41} \\ \hline
\end{tabular}
\end{table*}

\textbf{Computational complexity:} We evaluate computational complexity, parameter count, and inference speed using image inputs of $512 \times 512$ and map inputs of $256 \times 256$. As shown in Table.~\ref{tab:computational complexity}, despite incorporating the DINOv2 encoder, our OSMLoc-S model has fewer parameters, lower computational complexity, and slightly faster inference speed compared to OrienterNet. Even the larger OSMLoc-B model maintains real-time performance with nearly 10 FPS, which suggests that the performance improvement stems from the strong prior knowledge embedded in the foundation model rather than from a heavier or more complex architecture. To enable lightweight online deployment on edge devices, an alternative is to adopt efficient vision backbones such as MobileSAM~\citep{zhang2023faster} for extracting semantic- and geometry-aware features. Moreover, knowledge distillation~\citep{hinton2015distilling} also offers a promising direction to transfer performance from heavy teacher models to compact student networks.

\begin{figure}[ht]
\centering
\includegraphics[width=3.3in]{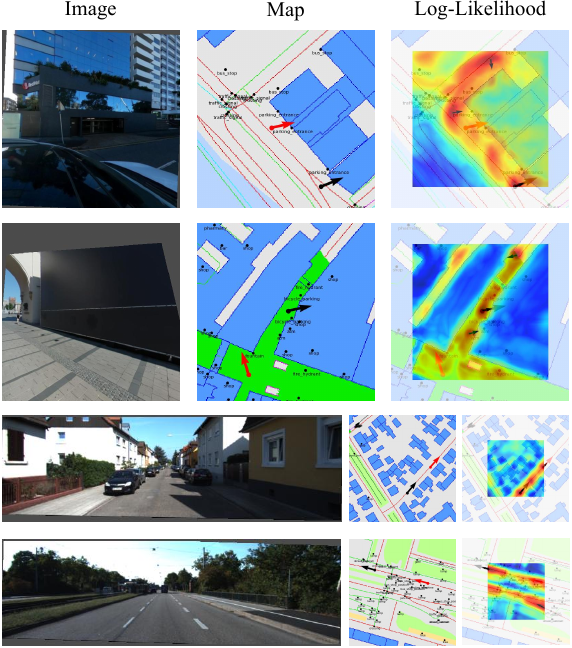}
\caption{Failure cases. The first and second ones are from the MGL dataset and the CC validation benchmark, while the third and last ones are from the KITTI dataset. Black arrow 
\( \boldsymbol{\textcolor{black}{\raisebox{-0.5ex}{\rotatebox{90}{$\rightarrow$}}}} \)
 represents the predicted pose and red arrow \( \boldsymbol{\textcolor{red}{\raisebox{-0.5ex}{\rotatebox{90}{$\rightarrow$}}}} \) represents the GT pose in the map.}
\label{fig:failure_case}
\vspace{-0.4cm}
\end{figure}

\textbf{Failure cases: }
The localization accuracy of our OSMLoc heavily depends on the quality, quantity and distinctiveness of the surrounding objects in the image. When visual information is limited, for example, when the camera’s field of view is obstructed by skyscrapers or dominated by monotonous surroundings, our method may struggle to estimate the pose accurately. Fig.~\ref{fig:failure_case} shows some failure cases under different conditions. In the first and second cases, the cameras are heavily occluded by tall nearby buildings, resulting in images that contain little more than the road and the facade of a single structure. Due to the repetitive visual patterns along the building-lined roadside, where similar scenes appear at multiple locations, the model cannot confidently pinpoint the query’s exact position, resulting in high localization ambiguity in the entire region. The third row, captured in an autonomous driving setting, presents a straight street bordered by visually repetitive buildings on both sides, a typical scene in many urban and suburban environments. This structural redundancy introduces significant uncertainty in determining the precise location. The last row shows a highway scene with few structured features or distinctive objects, resulting in high uncertainty along the vegetation-lined path.

\subsection{Application in sequential localization}\label{sec: SL}
From observed failures, we learn that while our method enhances single-image I2O localization performance, the task remains inherently ill-posed and uncertain due to limited FoV information and scale ambiguity, which may lead to occasional failures. For example, nearby crossings often appear similar and are difficult to distinguish, even for human users with a single glance. Leveraging sequential images could effectively mitigate these issues, and our method serves as a robust observation model for existing sequence-based approaches. 

Following previous approaches~\citep{chen2021range,zhou2021efficient}, we utilize the Monte Carlo Localization (MCL) framework with the particle filter to solve the sequential localization problem. For the first frame, we initialize $N_p$ particles onto the top-$N_p$ pose candidates and each particle represents a pose hypothesis $\boldsymbol{p}_1^i = (x_1^i,y_1^i,\theta_1^i)$. When the camera moves, the pose of each particle is updated based on the motion model with the control input $\mathbf{u}$. The expected observation from the predicted pose of each particle is compared with the actual observation acquired by the camera to update the weight of each sample. Particles are resampled based on the weight distribution and reduced for efficiency when the effective sample number falls below a predefined threshold. After several iterations, the particles gradually converge around the GT pose.

At timestamp $t$, MCL realizes a recursive Bayesian filter estimating the probability density $p(\boldsymbol{p}_t^i | \boldsymbol{S}_{1:t},\boldsymbol{u}_{1:t})$ over the pose $\boldsymbol{p}_t$ given all observation $\boldsymbol{S}_{1:t}$ and motion inputs $\boldsymbol{u}_{1:t}$ up to time $t$.  Here $\boldsymbol{S}$ is the matching score volume, and $t$ is reused as the timestamp for simplicity. The posterior is updated as:
\begin{align}
       p( & \boldsymbol{p}_t|\boldsymbol{S}_{1:t},\boldsymbol{u}_{1:t})= \eta p(\boldsymbol{S}_t | \boldsymbol{p}_t,\mathcal{M}) \\
      & \int p(\boldsymbol{p}_t | \boldsymbol{u}_t,\boldsymbol{p}_{t-1}) p(\boldsymbol{p}_{t-1} | \boldsymbol{S}_{1:t-1},\boldsymbol{u}_{1:t-1}) \, d\boldsymbol{p}_{t-1}, 
\end{align}
where $\eta$ is the normalized constant, $ p(\boldsymbol{p}_t | \boldsymbol{u}_t,\boldsymbol{p}_{t-1})$ is the motion model, and the $p(\boldsymbol{S}_t | \boldsymbol{p}_t,\mathcal{M}) $ is the observation model. We use the standard motion model in~\citep{thrun2002probabilistic} and focus on the observation model. For each particle $i$, we query the matching score $\boldsymbol{S}_t(x_t^i,y_t^i,\theta_t^i)$ from the score volume and approximate the likelihood   $p(\boldsymbol{S}_t | \boldsymbol{p}_t^i,\mathcal{M}) $  as a Gaussian observation model:
\begin{equation}
     p(\boldsymbol{S}_t | \boldsymbol{p}_t^i,\mathcal{M}) \propto w_t^i=\exp\big(\frac{\log \boldsymbol{S}_t(x_t^i,y_t^i,\theta_t^i)}{2\varsigma^2}\big) ,
\end{equation}
where $\varsigma$ controls the sensitivity of the observation. With the weight $w_t^i$ of each particle, the posterior pose $\hat{\boldsymbol{p}}_t$ is computed as the weighted average vector of the candidate poses:
\begin{equation}
    \hat{\boldsymbol{p}}_t= \frac{\sum_{i=1}^{N_p} w_t^i \boldsymbol{p}_t^i}{\sum_{i=1}^{N_p} w_t^i} .
\end{equation}

During implementation, the number of particles is set to $N_p=1000$ at initialization and reduced to 200 if converged. The sensitivity parameter $\varsigma$ is set to 2.0. The minimal and maximal sequence lengths are set to 3 and 10 respectively. To simulate the real motion pattern, random Gaussian noise $ \boldsymbol{\epsilon}_p$ is added to the GT pose. For the MGL dataset and CC validation benchmark, we sub-sample the sequences and ensure that frames are spaced by at least 4\text{m}, and $\epsilon_x, \epsilon_y \sim \mathcal{N}(0,0.5\text{m}), \epsilon_{\theta} \sim \mathcal{N}(0,0.1\pi)$. For the KITTI dataset, we use the raw sequences and $\epsilon_x, \epsilon_y \sim \mathcal{N}(0,0.05\text{m}), \epsilon_{\theta} \sim \mathcal{N}(0,0.01\pi)$.

\begin{figure}[ht]
\centering
\includegraphics[width=3.3in]{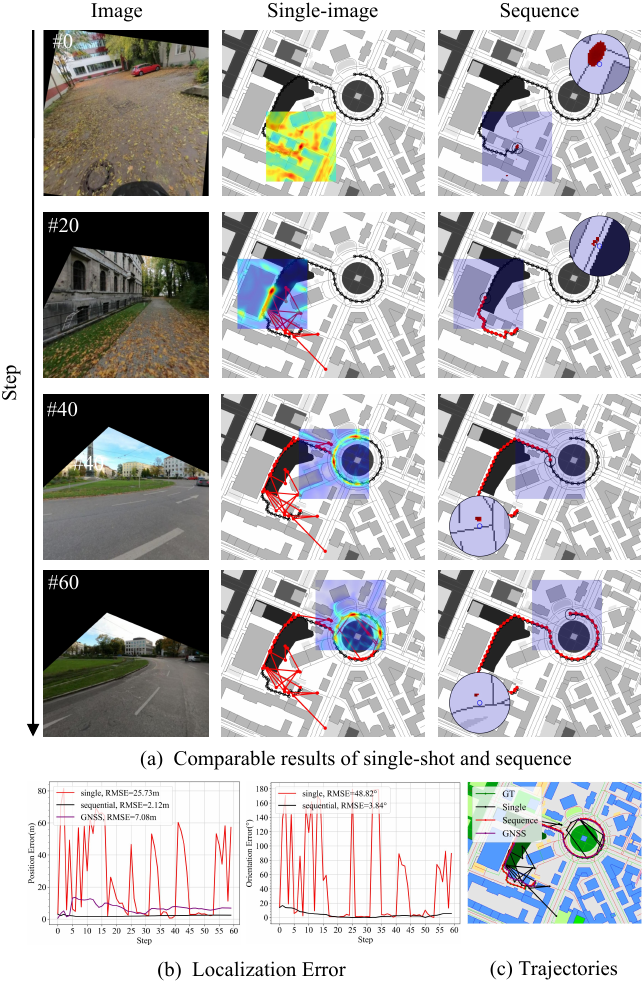}
\caption{Qualitative results of sequential localization in Munich. (a) comparable localization results of single-image and sequential images. (b) Incorporating multi-frame observation and the motion pattern reduces the ambiguity and consistently improves the accuracy and robustness. }
\label{fig:qua_seq_munich}
\end{figure}

The quantitative results of sequential localization on the MGL dataset, CC validation benchmark, and KITTI dataset are presented in Table.~\ref{table: seq_results} and Table.~\ref{tab:kitti_seq}. With the identical motion model and experimental settings, our method achieves higher position recall (PR) and orientation recall (OR) values than the baseline method. Compared with single-frame localization results, sequential localization significantly improves the accuracy and reduces ambiguity, as shown in  Fig.~\ref{fig:qua_seq_munich}.

\begin{table}[ht]
\footnotesize
\setlength{\tabcolsep}{2.5pt}
\caption{Sequential localization results on the MGL dataset and CC validation benchmark. We use \textbf{bold} font to indicate the best results.  "Detroit", "Munich", "Taipei" and "Brisbane" mean the city subsets of the CC validation benchmark. "single" means the single-image localization and "seq" means the sequential localization. "Ori" is short for OrienterNet.}
\label{table: seq_results}
\begin{tabular}{ccccccccc}
\hline
\multirow{2}{*}{Data} & \multirow{2}{*}{Setup} & \multirow{2}{*}{Method} & \multicolumn{3}{c}{PR @Xm$\uparrow$}                & \multicolumn{3}{c}{OR @X$^\circ$$\uparrow$}        \\ \cline{4-9} 
             &       &       & @1m            & @3m            & @5m            & @1$^\circ$             & @3$^\circ$             & @5$^\circ$            \\ \hline
 \multirow{4}{*}{\rotatebox{90}{MGL}}   & \multirow{2}{*}{single} & Ori        &       10.69          & 42.12          & 54.01          & 19.22          & 48.56          & 63.59          \\
 & & Ours           & 15.30 & 46.05 & 59.40 & 21.79 & 56.57 & 71.29 \\ \cline{2-9}
 & \multirow{2}{*}{seq} &  Ori       & 21.47          & 59.56          & 72.90          & 22.32          & 56.70          & 71.47     \\
  &      &  Ours   & \textbf{25.13} & \textbf{61.74} & \textbf{77.42} & \textbf{25.77} & \textbf{61.64} & \textbf{75.88}   \\ \hline
\multirow{4}{*}{\rotatebox{90}{Detroit}} & \multirow{2}{*}{single} & Ori               & 7.64           & 33.11          & 46.24          & 13.66          & 32.10          & 44.57          \\
& & Ours               & 13.37 & 51.79 & 63.01 & 19.75 & 48.57 & 62.41 \\ \cline{2-9}
& \multirow{2}{*}{seq}  & Ori               & 15.41          & 53.60          & 66.13          & 16.79          & 42.39          & 57.01          \\
                         &  & Ours             & \textbf{23.74} & \textbf{71.16} & \textbf{80.04} & \textbf{23.56} & \textbf{59.71} & \textbf{74.22}
                          \\ \hline
\multirow{4}{*}{\rotatebox{90}{Munich}}     & \multirow{2}{*}{single} & Ori               & 2.68           & 17.90          & 35.11          & 12.48          & 32.75           & 47.40         \\
& & Ours   &             4.36  & 22.03 & 40.58 & 15.35 & 41.79  & 59.65          \\ \cline{2-9}
& \multirow{2}{*}{seq}            & Ori               & 4.14           & 31.92           & 54.41          & 15.99          & 43.45          & 58.78          \\
                           & & Ours             & \textbf{7.99}  & \textbf{40.57} & \textbf{63.10} & \textbf{21.19} & \textbf{53.07} & \textbf{71.79} \\ \hline
 \multirow{4}{*}{\rotatebox{90}{Taipei}}  & \multirow{2}{*}{single} & Ori              & 1.23           & 8.17           & 16.13          & 5.88           & 16.40         & 24.79 \\
 & & Ours              & 1.44  & 9.72 & 19.18 & 7.32  & 21.31 & 32.10          \\ \cline{2-9}
 & \multirow{2}{*}{seq} & Ori               & 2.62           & 15.36           & 26.18          & 5.30           & 16.54          & 25.80          \\
                           & & Ours             & \textbf{2.74}  & \textbf{16.10} & \textbf{29.60} & \textbf{6.37} & \textbf{21.31} & \textbf{32.82} \\ \hline
 \multirow{4}{*}{\rotatebox{90}{Brisbane}}  &  \multirow{2}{*}{single} & Ori               & 0.21           & 2.60           & 7.47           & 4.66           & 9.73          & 17.60          \\
 & & Ours               & 0.27  & 2.88  & 9.18  & 5.00  & 15.34 & 24.86          \\ \cline{2-9}
 &     \multirow{2}{*}{seq}         & Ori               & 1.03  & 7.16           & 16.39          & 4.41           & 12.74          & 20.59          \\
                          & & Ours             & \textbf{1.38}  & \textbf{9.16} & \textbf{17.70} & \textbf{6.54} & \textbf{16.74} & \textbf{26.93} \\ \hline
\end{tabular}
\end{table}

\begin{table*}[t]
\caption{Sequential localization results on the KITTI dataset. We use \textbf{bold}
font to indicate the best results. "single" means the single-image localization and "seq" means the sequential localization.}
\label{tab:kitti_seq}
\footnotesize
\setlength{\tabcolsep}{10.0pt}
\begin{tabular}{ccccccccccc}
\hline
\multirow{2}{*}{Method} &  \multirow{2}{*}{Setup}   & \multicolumn{3}{c}{LatR @Xm$\uparrow$}                                                                                      & \multicolumn{3}{c}{LonR @Xm$\uparrow$}                                                                                   & \multicolumn{3}{c}{OR @X$^\circ$$\uparrow$}                                                                                       \\ \cline{3-11}                        
& &  @1m                                & @3m                                & @5m                                & @1m                                & @3m                                & @5m                                & @1$^\circ$                                 & @3$^\circ$                                 & @5$^\circ$                                 \\ \hline
 OrienterNet        &  \multirow{2}{*}{single}   & 62.03          & 90.89          & 95.73          & 19.99          & 51.60          & 65.78          & 29.93          & 72.81          & 87.79          \\ 
Ours  &    & 66.60 & 95.05 & 97.41 & 29.40 & 63.86 & 74.52 &  35.57 & 78.28 & 92.96 \\   \hline
OrienterNet & \multirow{2}{*}{seq}   & 66.86          & 93.39          & 97.16          & 22.84          & 59.79          & 75.50          & 45.75          & 87.55          & 95.43          \\ 
Ours     &  &    \textbf{75.02} & \textbf{96.43} & \textbf{98.41} & \textbf{35.75} & \textbf{73.23} & \textbf{84.41} & \textbf{50.83} & \textbf{90.51} & \textbf{97.57} \\
                           \hline
\end{tabular}
\end{table*}

\section{Conclusion}
In this paper, we present the OSMLoc, a brain-inspired single image-to-OpenStreetMap (I2O) visual localization framework. The proposed method fuses the recent foundation model Depth Anything for geometric guidance, and fully exploits semantic information from the OSM data to enhance the perception capability of the image model. A novel cross-area and cross-condition (CC) validation benchmark is proposed to evaluate the accuracy, robustness and generalizability of I2O visual localization methods. Experiments on the MGL dataset and CC validation benchmark have demonstrated that our OSMLoc achieves higher localization accuracy, while extensive experiments on the KITTI dataset highlight its application value in the autonomous driving area. Sequential localization experiments underscore the potential practical value. We hope that our OSMLoc will benefit the relevant communities in the field.

\nolinenumbers

\section*{CRediT authorship contribution statement}
\textbf{Youqi Liao:} Writing - original draft, Methodology, Data curation. \textbf{Xieyuanli Chen:} Writing – review \& editing, Methodology, Conceptualization. \textbf{Shuhao Kang:} Writing - review \& editing, Visualization.  \textbf{Jianping Li:} Writing - review \& editing, Supervision, Funding acquisition. \textbf{Zhen Dong:} Writing - review \& editing, Supervision.  \textbf{Hongchao Fan:} Writing - review \& editing, Validation. \textbf{Bisheng Yang}: Supervision, Funding acquisition.

\section*{Acknowledgment}
This study was jointly supported by the National Natural Science Foundation Project (No.42201477, No.42130105).

\bibliographystyle{model1-num-names}

\bibliography{cas-refs}

\end{document}